\documentclass[letterpaper, 10 pt, conference]{ieeeconf}  
\IEEEoverridecommandlockouts
\usepackage{graphicx}
\usepackage{mathptmx} 
\usepackage{times} 
\usepackage{amsmath} 
\usepackage{amssymb}  
\usepackage{units}
\usepackage{verbatim}  
\usepackage{nicefrac}		
\usepackage{acronym}		
\usepackage{glossaries}		
\usepackage[usenames,dvipsnames]{xcolor}			
\usepackage{cite} 
\usepackage{url} 
\usepackage[multidot]{grffile} 
\usepackage{multirow}
\usepackage{booktabs}
\makeatletter
\let\NAT@parse\undefined
\makeatother
\usepackage{hyperref}
\hypersetup{colorlinks,allcolors=black}
\newacronym{ACFR}{ACFR}{Australian Centre for Field Robotics}
\newacronym{USyd}{USyd}{the University of Sydney}
\newacronym{AUV}{AUV}{autonomous underwater vehicle}
\newacronym{UAV}{UAV}{unmanned aerial vehicle}
\newacronym{SLAM}{SLAM}{simultaneous localisation and mapping}
\newacronym{SfM}{SfM}{structure-from-motion}
\newacronym{SNR}{SNR}{signal-to-noise ratio}
\newacronym{DFT}{DFT}{discrete Fourier transform}
\newacronym{FFT}{FFT}{fast Fourier transform}
\newacronym{SIFT}{SIFT}{scale invariant feature transform}
\newacronym{TP}{TP}{true positive}
\newacronym{FP}{FP}{false positive}
\newacronym{BuFF}{BuFF}{burst feature finder}
\newacronym{NIR}{NIR}{near-infrared}
\newacronym{SURF}{SURF}{speeded up robust features}
\newacronym{DoG}{DoG}{difference of Gaussians}
\newacronym{LiFF}{LiFF}{light field features}
\newacronym{ROC}{ROC}{receiver operating characteristic curve}
\newacronym{EV}{EV}{exposure value}
\newacronym{DOF}{DOF}{degree-of-freedom}
\newacronym{ORB}{ORB}{oriented FAST and rotated BRIEF}
\newacronym{R2D2}{R2D2}{repeatable and reliable detector and descriptor}
\newacronym{RANSAC}{RANSAC}{random sample consensus}

\title{\LARGE \bf LBurst: Learning-Based Robotic Burst Feature Extraction for 3D Reconstruction in Low Light}
\author{Ahalya Ravendran$^{1}$, Mitch Bryson$^{2}$, Donald G. Dansereau$^{2}$%
\thanks{$^{1}$Ahalya Ravendran is with the Imaging and Computer Vision Research Group, Commonwealth Scientific and Industrial Research Organisation (CSIRO), Sydney, Australia
{\tt\footnotesize ahalya.ravendran@data61.csiro.au}}%
\thanks{$^{2}$Donald G. Dansereau and Mitch Bryson are with the Australian Centre for Robotics (ACFR), School of Aerospace, Mechanical and Mechatronic Engineering, The University of Sydney, 2006, NSW, Australia.
{\tt\footnotesize mitch.bryson, donald.dansereau@sydney.edu.au}}%
}

\newcommand{\myquat}[1]{\bar{#1}}

\newcommand{\q}{\myquat{q}}

\newcommand{\Cq}[2]{\boldsymbol{C}(\q)}

\begin{document}
\maketitle
\thispagestyle{empty}
\pagestyle{empty}

\begin{abstract}
Drones have revolutionized the fields of aerial imaging, mapping, and disaster recovery. However, the deployment of drones in low-light conditions is constrained by the image quality produced by their on-board cameras. In this paper, we present a learning architecture for improving 3D reconstructions in low-light conditions by finding features in a burst. Our approach enhances visual reconstruction by detecting and describing high quality true features and less spurious features in low signal-to-noise ratio images. We demonstrate that our method is capable of handling challenging scenes in millilux illumination, making it a significant step towards drones operating at night and in extremely low-light applications such as underground mining and search and rescue operations.
\end{abstract}

\section{Introduction}
\label{sec:intro}

The ubiquity and accessibility of drones has led to their widespread use in various applications, including aerial photography~\cite{mehta2021domain}, mapping~\cite{zhu2021revisiting}, surveillance~\cite{ashraf2021dogfight}, monitoring~\cite{asanomi2023multi}, and rescue missions~\cite{zhu2021msnet}. However, operating them at night remains a significant challenge due to imaging limitations in low signal conditions. While drones can carry their own light sources and additional sensors, it is not always practical due to weight or power limitations, and undesirable in applications such as surveillance and nocturnal wildlife studies. Thus, there is a need for an imaging approach using conventional cameras for reconstruction in low-light conditions.

\begin{figure}[t]
    \centering
    \includegraphics[width=\linewidth]{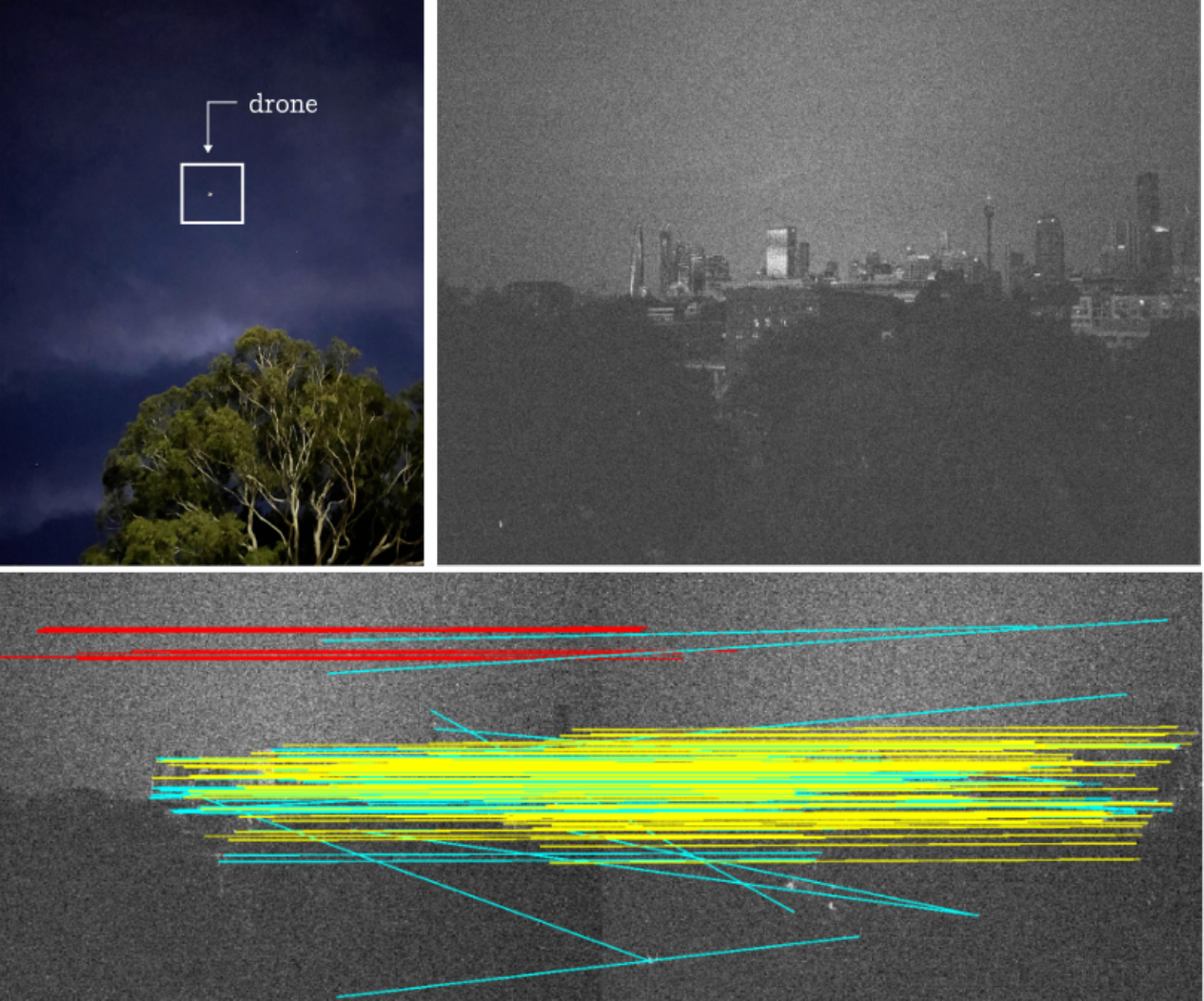}
    \caption{Feature matching at night: A commercial drone (top-left) captures imagery (top-right) in millilux illumination at night. This imagery is too noisy for conventional 3D reconstruction due to the high rate of spurious features detected by conventional methods like SIFT (bottom, blue) and less reliable matches provided by learned feature extractors like R2D2 (bottom, red) on noisy images. The proposed LBurst yields higher-quality true features from reliable regions of the images (bottom, yellow) for reconstruction. In this work, we demonstrate that our feature finder facilitates 3D reconstruction in millilux illumination for existing commercial drones.}
    \label{fig_abstract}
\end{figure}

Burst imaging, a recently introduced technique for low-light mobile photography, involves capturing a sequence of images with a large overlap between frames over a short exposure time. This process generates a single image with higher signal-to-noise ratio (SNR)~\cite{hasinoff2016burst}. Adapting burst imaging for robotics, burst-based structure-from-motion~(SfM)~\cite{aR2021} established the viability of using robotic burst imaging for reconstruction in low light. However, detecting features from burst-merged images tends to produce misalignment errors~\cite{aR2021}. 

In \cite{aR2022}, the authors presented a physics-based feature extraction approach, burst feature finder (BuFF), to directly find features in a robotic burst demonstrating the success of burst-based approaches for reconstruction in low light. Despite achieving comparable performance to single image-based learning approaches \cite{detone2018superpoint,revaud2019r2d2} for feature extraction, learning-based methods for burst feature extraction in low-light conditions remain relatively less explored.

In this paper, we present a learning-based pipeline for robotic burst imaging that utilizes joint detection and description to locate keypoints within a burst. By leveraging metric learning loss from existing 2D feature extractors \cite{revaud2019r2d2,he2018local} and adapting for robotic burst, we identify robust features for light-constrained reconstruction.

Our key contributions are:
\begin{itemize}
\item We introduce a learning-based robotic burst feature finder, LBurst, a joint feature detector and descriptor that finds learned features with well defined scale within a low-light robotic burst,
\item We enable robots to detect noise-tolerant features in low light by employing consistent motion within bursts and leveraging uncorrelated noise between them, and
\item We demonstrate overall improved 3D reconstruction from drone images in millilux conditions, outperforming single image-based feature extractors and robotic burst feature finder in low-SNR scenes.
\end{itemize}

We evaluate our approach with the single-image-based repeatable and reliable detector and descriptor (R2D2)~\cite{revaud2019r2d2} to highlight the benefits of burst imagery, as it allows direct comparison to single-image approaches. We employ HPatches dataset~\cite{balntas2017hpatches} with varying illuminations and viewpoints to generate synthetic robotic bursts. Our burst-based method shows improved feature matching and camera pose estimates using COLMAP~\cite{schonberger2016structure}.

To further validate, we use captured burst imagery from DJI Mini Pro 3 and Phantom Pro 4 drones operating in low-light conditions. We compare our proposed approach against 2D feature extraction methods on conventional drone images and against other robotic burst imaging schemes. We demonstrate high quality true features, leading to better quality matches for scenes captured in millilux illumination as shown in \autoref{fig_abstract}. We show an overall improvement in reconstruction performance, resulting in more complete 3D models. The code and dataset will be publicly available upon paper acceptance. This work aims to identify learning-based features in low-light robotic bursts that enhance low-light reconstruction, benefiting a wide range of drone applications, including drone delivery at night and search and rescue operations. 
\section{Related Studies}
\label{sec:related}

Burst imaging has emerged as a promising solution for enhancing low-light mobile photography~\cite{hasinoff2016burst}. While current research focuses on generating appealing single burst images by addressing handshake motion and noise commonly encountered in handheld mobile photography~\cite{hasinoff2016burst, liba2019handheld, wronski2019handheld, mildenhall2018burst}, these methods lack direct applicability for robotic applications. This is because drones operate in lower light conditions compared to hand-held burst photography with different motion profiles \cite{aR2022} to generate reliable estimates for downstream tasks such as 3D reconstruction. 

Ravendran et al. \cite{aR2021} present a robotic burst imaging method capturing challenging low-light burst imagery with drone-specific motion profiles, showcasing improved 3D reconstruction under low-light conditions. Addressing the pairwise misalignment associated with this method, a physics-based burst feature finder (BuFF) \cite{aR2022} directly identifies 2D + time features in a robotic burst through local extrema search in a scale-apparent motion search space. In our work, we introduce an end-to-end learning-based approach, offering improved feature performance in low light with the potential for real-time operation.

SuperPoint \cite{detone2018superpoint} computes feature locations and descriptors simultaneously through pseudo-ground truth generation for repeatable features, while discrete keypoints (DISK) \cite{tyszkiewicz2020disk} uses a probabilistic approach based on ground-truth poses and depth. However, in low-light drone imagery, ground truth is unavailable due to noise and generate inaccurate depth and pose estimates.

Recent end-to-end methods \cite{han2015matchnet,he2018local,revaud2019r2d2, yi2016lift, dusmanu2019d2net} outperform earlier deep learning approaches in image matching and long-term visual localization \cite{dusmanu2019d2net}, especially in challenging conditions like low-light, where traditional methods struggle with spurious features \cite{lowe2004distinctive, Bay2006, Rublee2011}. These methods use a single network for both keypoint detection and description, but most rely on convolutional neural networks (CNNs) \cite{lecun1995convolutional}, limiting their receptive field due to the inherent locality of CNNs. 

R2D2 addresses ambiguities in repetitive regions but still struggles with low SNR images as it lacks broader visual context \cite{wang2023attention}. Despite these limitations, R2D2 remains state-of-the-art for CNN-based feature extraction in visually challenging conditions. While DISK outperforms R2D2 through probabilistic reinforcement learning, it requires ground truth poses and depth, which are often unavailable in drone imagery. Therefore, we selected R2D2 as the foundation for developing our burst-based feature extraction approach, though this method is adaptable to other single-image feature extractors as well.
\section{Learned Robotic Burst Feature Extraction}
\label{sec:lburst}

We locate learned features with well-defined scale within low-light robotic bursts using a self-supervised approach enforcing burst pair similarity as described in \autoref{architecture}. By employing a flow map using known transformation between common images (i.e., the middle frames) of the ground truth bursts during training, the network learns to find noise-tolerant features. These features are compact 128-dimensional descriptors, obtained using mean average-precision (mAP) similar to R2D2 \cite{revaud2019r2d2} and descriptors optimized for average precision (DOAP) \cite{he2018local}. We select keypoints based on detection and description scores, applied across scales to identify multi-scale features. During inference, we compute multi-scale feature locations and descriptors within a single burst, representing the pose of the common image. For reconstruction, this process extends over multiple bursts captured along a drone's trajectory. The following outlines the joint architecture for detection and description and the burst generation pipeline for training and inference.

\begin{figure}[t]
    \centering
    \includegraphics[width=\linewidth]{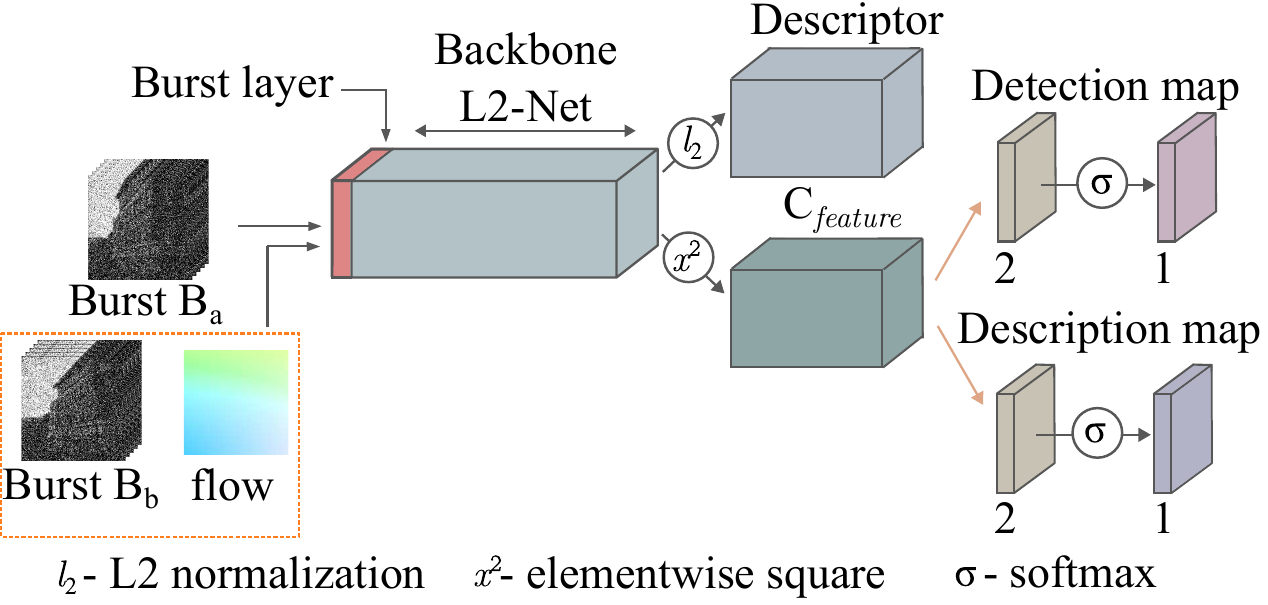}
    \caption{LBurst architecture employs two bursts of images and a flow map with the known ground-truth transformations between the common images of a pair of robotic bursts during training. A specialized burst layer handles burst image processing. The architecture employs a fully convolutional L2-Net, similar to R2D2, generating a 128D descriptor with L2 normalization and a squared element-wise operation yielding ${\mathit{C}_\mathit{feature}}$. Following a 1 x 1 convolution for dimension reduction, a softmax operation computes associated confidence maps.}
    \label{fig_architecture}
\end{figure}

\subsection{Joint Learning of Detection and Description}
\label{architecture}
We employ a fully-convolutional network with L2-Net \cite{tian2017l2} as the backbone architecture to predict robust features for a burst $B$ of size ${H}\times{W}\times{N}$, where ${H}\times{W}$ represents the size of each image, and $N$ denotes the number of images within the burst. The network generates a 3D tensor $F\in\mathbb{R}^{{H}\times{W}\times{D}}$ that corresponds to a set of dense $D$-dimensional descriptors, one per pixel of the common image. Because feature-based SfM can handle large occlusions between the bursts, we focus on finding features of the common image that represents the burst, i.e., the middle frame in the case of a robotic burst \cite{aR2021}. The network additionally generates a detection confidence map $K\in[0, 1]^{{H}\times{W}}$ with local maxima corresponding to the keypoint locations of the common image and a confidence map $R\in[0, 1]^{{H}\times{W}}$ that indicates the estimated reliability of the descriptor at each pixel.

We train our model using robotic bursts paired with flow maps, facilitating a known transformation between common burst images as described in \autoref{burst_generate}. Our adaptable burst layer handles multiple input images, maintaining spatial resolution through dilated convolutions. For faster inference, we replace the ${8\times8}$ final convolution layer of L2-Net with three successive ${2\times2}$ layers, similar to R2D2. In contrast to R2D2, our model incorporates flow maps, imposing spatial pose constraints to capture temporal relationships within burst images.

We generate the detection confidence map $K$ as a self-supervised task by maximizing the cosine similarity between the pairs of patches from the common images of bursts $B_a$ and $B_b$. We define a set of overlapping patches of size ${M}\times{M}$ and enforce local maxima by both averaging the cosine similarity loss $L_c$ over the patches and incorporating a local peakiness loss $L_p$ inspired by \cite{revaud2019r2d2}. Because the noise in low SNR images are random in both spatial and temporal domains, locating features based on similarity avoids spurious features. This loss computation promotes noise tolerant features within a low-light robotic burst and consistency of features between the bursts. The overall detection loss $\mathit{L}$ in a robotic burst $B_a$ is computed as
\begin{equation} \label{eq:detection_loss} L(B_a,B_b,U_i) = L_c(B_a,B_b,U_i) + \frac{L_p(B_a) + L_p(B_b)}{2}, \end{equation}
where $B_b$ represents bursts of images transformed by the flow map $U_i$, ensuring alignment between the common images of $B_b$ and $B_a$.

We aim to maximize the average-precision (AP) to evaluate the reliability of the patches in the description confidence maps $R$. We extract dense local descriptors $F$ and predict for each descriptor $F_ij\in\mathbb{R}^D$, a reliability score $R$. This is done by computing the Euclidean distance matrix between all patch descriptors in the batch, comparing patches from a common image in burst $B_a$ with those in burst $B_b$. By training with interpretable threshold, we exclude non-distinctive regions during reconstruction, further reducing the presence of unreliable features that have lower matching probabilities.

\subsection{Robotic Burst Generation}
\label{burst_generate}
\begin{figure}[b]
    \centering
    \includegraphics[width=\linewidth]{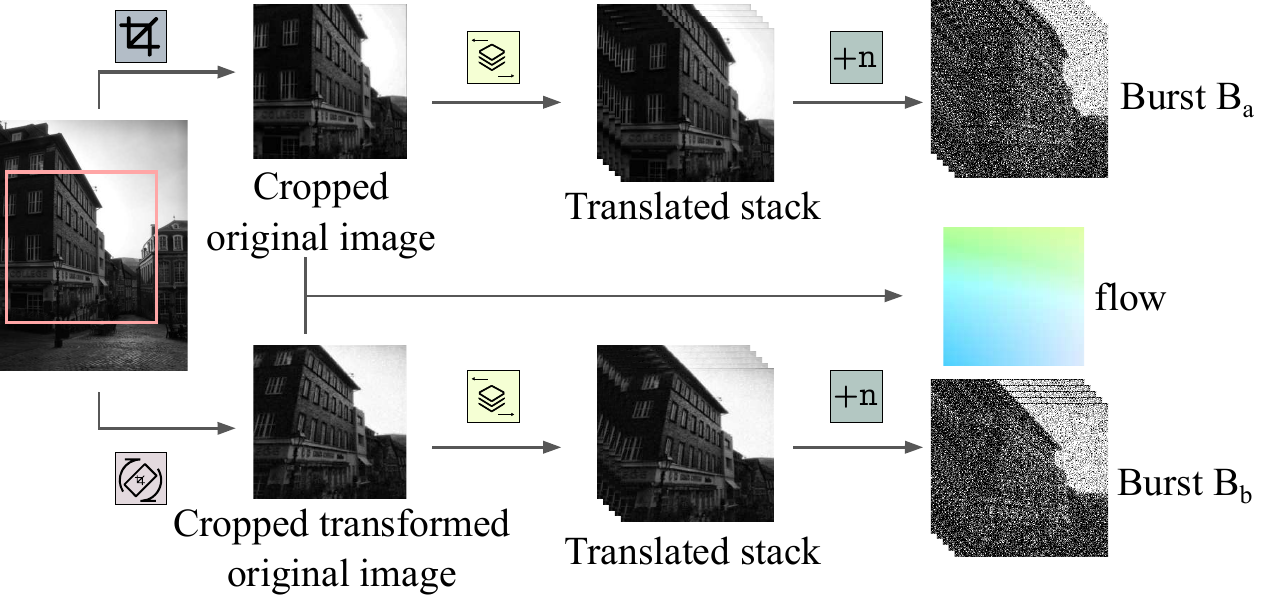}
    \caption{Overview of the synthetic robotic burst generation for model training. Cropping and transforming a single image to yield a pair of perspective-altered scenes, and by applying uniform 2D translations within each burst we generate a pair of burst images. We introduce random Gaussian noise to each image in a burst and employ flow maps with known transformations to ensure an accurate vector field between common images within the generated bursts.}
    \label{fig_burstgenerate}
\end{figure}

For training, we generate a pair of robotic bursts from a single image, applying homographic transformations to simulate different camera angles while keeping the scene consistent as in \autoref{fig_burstgenerate}. Each burst undergoes uniform 2D translations, with additional random translations between bursts. We also introduce Gaussian noise to simulate thermal noise in low-light conditions. The common image between bursts is the middle image, minimizing motion variation within the burst compared to using the first or last image, which would increase apparent motion. A flow map, $U_i$, is generated to establish ground truth spatial correspondence between the bursts, disregarding intra-burst motion.

Our architecture learns to handle various challenges encountered on robotic bursts during inference, such as local occlusions, parallax motion, and warp artifacts even though the model is not explicitly trained on these specific attributes. This is because we use local cosine similarity over small patches for feature detection, and average them over all the patches and by doing so, we treat apparent motion associated with the patches locally.
\section{Results}
\label{sec:results}
\subsection{Implementation Details}
\label{training}
Our learning-based burst feature finder, LBurst, identifies sparse feature locations and descriptors across multi-scale bursts, scoring them based on detection and description confidence for robustness in low light. We evaluate LBurst on synthetic HPatches robotic burst dataset and demonstrate real-world 3D reconstruction in millilux illumination using drone-captured bursts.

\subsubsection{Training Datasets}
\label{datasets}
We train our method using two single image-based datasets commonly used for feature extraction methods. We incorporate 4479 images from the Aachen dataset \cite{sattler2018benchmarking}, and 3190 random web images \cite{radenovic2018revisiting} as original images in constructing 5-frame robotic bursts. We exclude the captured robotic bursts during training and show that our model generalizes well to drone-captured data featuring urban cityscapes, including buildings, streets and trees.

\subsubsection{Training Parameters}
\label{parameters}
We generate 5-frame low-light robotic bursts following \autoref{burst_generate} with a maximum translation of 30 pixels and noise variance between 0.3 and 0.6 on normalized images. The model is trained for 25 epochs using the the Adam optimizer \cite{kingma2014adam} with a fixed learning rate of 0.0001 and a weight decay of 0.0005.

\subsection{Rendered Imagery}
\label{synthetic}
We evaluate our adaptation of R2D2 for low-light robotic bursts compared to the original R2D2 on noisy 2D images. We generate bursts from the HPatches dataset \cite{balntas2017hpatches} comprising of progressively challenging images in terms of viewpoint and illumination. We use each image in the dataset as an original image to generate a robotic burst with uniform translation and random noise as described in \autoref{burst_generate}. For camera pose estimation, we focus on bursts with varying viewpoints across 56 HPatches scenes.

\subsubsection{Matching Accuracy in Synthetic Noise}
\label{sec:matchingHPatches}
\begin{figure}[t]
    \centering
    \includegraphics[width=\linewidth]{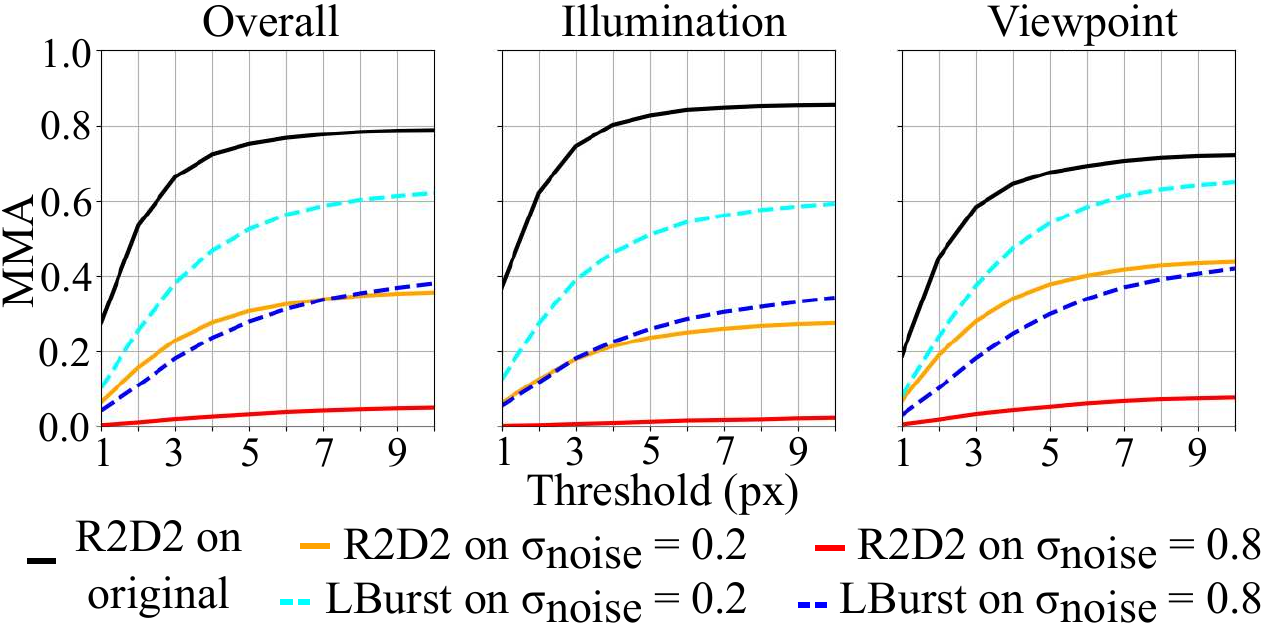}
    \caption{Mean matching performance for scenes with varying viewpoint and illumination. Operating R2D2 on the original HPatches dataset with high SNR (black) acts as a baseline. Operating R2D2 on the generated HPatches robotic burst common images fails to produce accurate matches for moderate (orange) and strong (red) noise. Our method outperforms R2D2 in producing improved feature matches in both moderate (cyan) and strong (blue) noise.}
    \label{fig_noisematch}
\end{figure}

\begin{table}[b]
    \centering
    \caption{The mean matching accuracy and repeatability score for all bursts in the generated HPatches robotic burst dataset. Bold highlights the overall best results.}
    \begin{tabular}{@{}ccc@{}}
    \toprule
    Method        & MMA@3  & Repeatability \\
    \midrule
    Gold Standard   & 0.68 & 0.31       \\
    \midrule
    R2D2  & 0.25 & 0.10        \\
    LBurst (Ours) & \textbf{0.44} & \textbf{0.29} \\
    \bottomrule
    \end{tabular}
    \label{table_HPatchesFeat}
\end{table}

Matching accuracy demonstrates the ability of a model to correctly match pixel correspondences between images of the same scene. We compare the mean matching accuracy (MMA) for various pixel error thresholds \cite{dusmanu2019d2net} using our feature finder on generated low-light robotic bursts and single-image based R2D2 on corresponding common images for all 108 scenes as in \autoref{fig_noisematch}. Ground truth is computed by running R2D2 on high-SNR images (black). Our method outperforms R2D2 in matching performance at all thresholds under moderate and strong noise, though performance declines with increasing noise.

We also evaluate repeatability score, measuring the number of putative matches for detected keypoints under strong noise as shown in \autoref{table_HPatchesFeat}. Our method outperforms R2D2, showing improved matching quality with higher match counts, further demonstrating the benefit of bursts in low-light conditions.

\subsubsection{Camera Trajectory Accuracy in Synthetic Noise}
\label{sec:camera}
We evaluate the accuracy of our camera trajectory estimation by validating the camera poses estimated using COLMAP \cite{schonberger2016structure}, aligning and scaling them with ground truth due to the inherent scale ambiguity in monocular SfM. The scale factor is determined by the distance between the initial registered images. We compute absolute and relative pose errors for translation and rotation as shown in~\autoref{table_camera}. Our method successfully converges in 60\% of diverse viewpoint scenes, outperforming R2D2 (12.5\%) with accurate pose estimates on the noisy HPatches robotic burst dataset.

\begin{table}[t]
    \centering
    \caption{The mean translation error in distance units and the mean rotational error in degrees for the reconstructed camera pose estimates of the generated HPatches viewpoint subset bursts at strong noise. Bold highlights the overall best results.}
    \label{table_camera}
    \resizebox{0.9\linewidth}{!}{
        \begin{tabular}{ccccc}
            \toprule
            Method & \multicolumn{2}{c}{Absolute trajectory error} & \multicolumn{2}{c}{Relative pose error} \\
             & trans. & rot. & trans. & rot. \\ 
            \midrule
            R2D2 & 3.30 & 2.18 & 5.18 & 0.86 \\
            LBurst & \textbf{1.32} & \textbf{1.03} & \textbf{1.90} & \textbf{0.45} \\
            \bottomrule
        \end{tabular}
    }
\end{table}

\subsection{Drone-based Validation}
\label{sec:dronevalidation}

\begin{figure}[b]
    \centering
    \includegraphics[width=\linewidth]{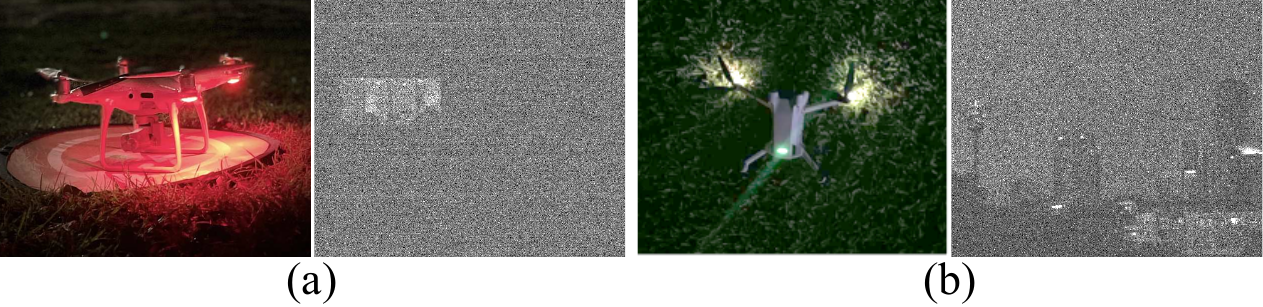}
    \caption{Low-light scene capture using two drones. (a) (Left) DJI Phantom 4 Pro used for burst capture at 0.125 ms exposure in millilux illumination; (right) Common image from the captured burst. (b) (Left) DJI Mini 3 Pro used for burst capture at the same conditions; (right) Common image from the captured burst. The complete dataset comprises 10 scenes with 15 bursts of 5 images each.}
    \label{fig_drone}
\end{figure}

We validate the learned burst features for low-light 3D reconstruction by deploying drones, a DJI Mini Pro 3 and a DJI Phantom Pro 4 in millilux illumination as shown in \autoref{fig_drone}. During drone flights, we recorded multiple trajectories, capturing 10 different scenes including buildings, trees, and streets. For each scene, we collected 15 bursts, with each burst consisting of 5 images. The bursts exhibit approximately linear 2D motion within each sequence.

\begin{figure}[t]
    \centering
    \includegraphics[width=\linewidth]{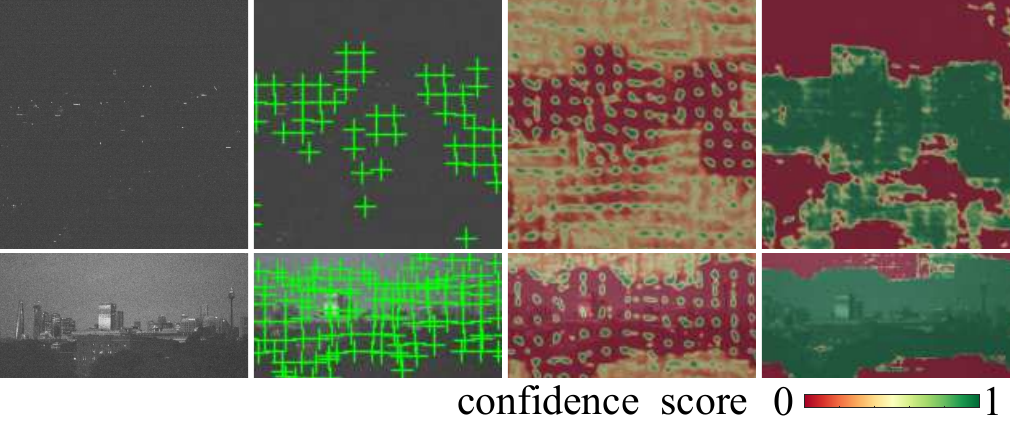}
    \caption{Keypoint localization with LBurst on captured imagery from two commercial drones. (Top row) DJI Phantom 4 Pro captures strong noise low-light scenes at 0.02 - 0.08 lux millilux conditions. (Bottom row) DJI Mini 3 Pro captures moderate noise low-light scenes at 0.05 - 0.12 lux millilux conditions. For both robotic bursts, we show from left to right, top scored keypoints, detection confidence map, and descriptor confidence maps using a color scheme where red indicates lower confidence and green signifies higher confidence.}
    \label{fig_features}
\end{figure}

Both of our commercial drones have monocular cameras. The DJI Mini 3 Pro camera has an f/1.7 lens which captures 10 bit depth images (4032 x 2268) in 0.125 seconds at ISO 6400. Meanwhile, the DJI Phantom Pro 4 camera has an f/2.8 lens and captures 16-bit images (4864 x 3648) similarly in 0.125 seconds, but at ISO 12800. We capture burst images under varying millilux illumination conditions, using a light meter for intensity measurement. Although both cameras were set to equal exposure values, the DJI Phantom drone's images exhibit stronger noise due to extreme low-light conditions and reduced natural light. Therefore, no ground truth was available for accurate comparison in these challenging scenarios.

\begin{figure}[b]
    \centering
    \includegraphics[width=\linewidth]{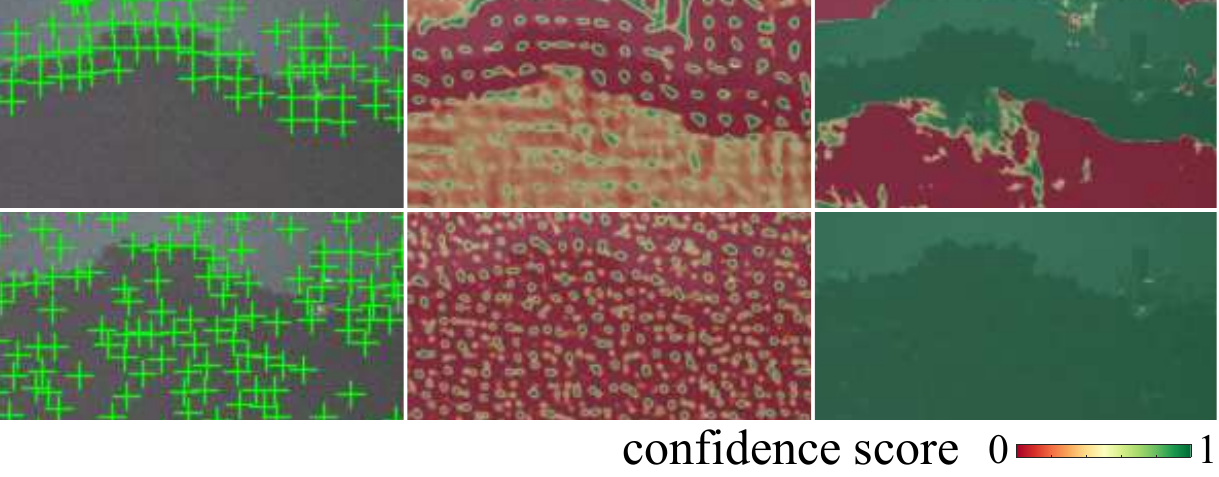}
    \caption{Comparison of our method LBurst (top row) with R2D2 (bottom row) on a low-light scene.  (Left) LBurst demonstrates high quality true features and fewer spurious features compared to R2D2. (Middle) LBurst detection confidence map highlights repeatable regions as local extrema while R2D2 lacks such consistency. (Right). Our descriptor confidence map targets excluding sky areas for improved reconstruction, unlike R2D2 which considers all regions reliable at the presence of strong noise.}
    \label{fig_LB-R2D2}
\end{figure}

\subsubsection{Feature Extraction Performance}

\begin{figure}
    \centering
    \includegraphics[width=\linewidth]{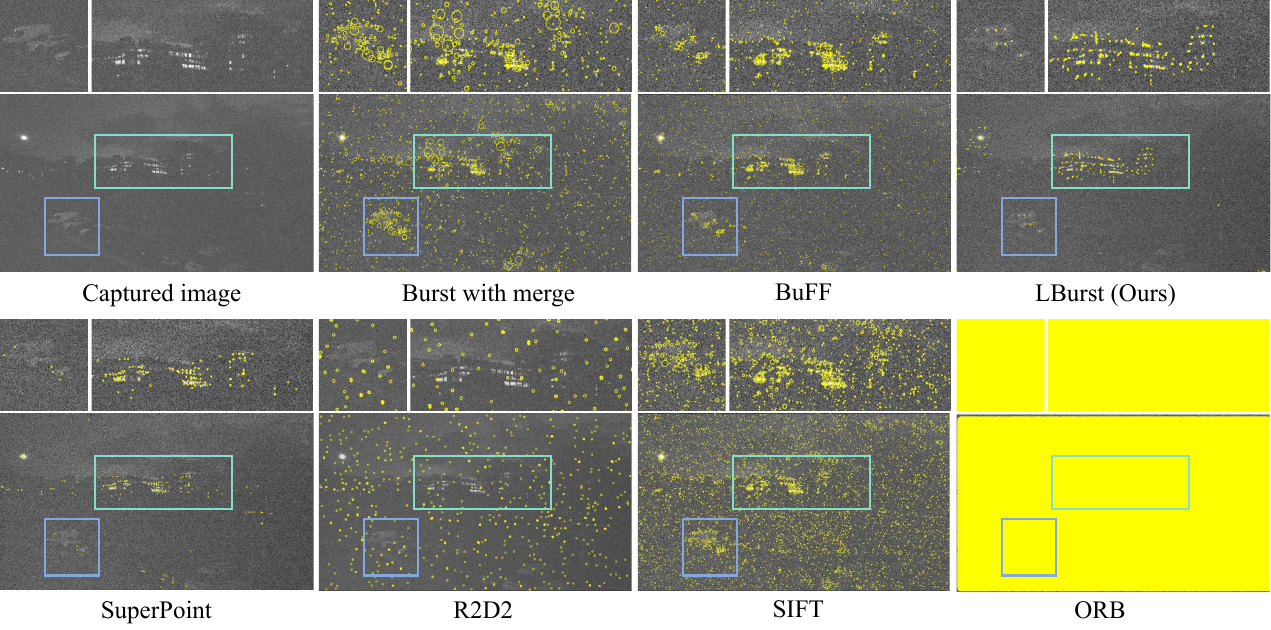}
    \caption{Comparison of feature performance on a captured scene (top, left) using burst-based techniques (top row) and single image-based techniques (bottom row), highlighting features associated with background (green) and foreground (blue) regions.}
    \label{fig_allfeatures}
\end{figure}

We demonstrate LBurst features and confidence maps for both detection and descriptors on scenes captured with the DJI Phantom 4 Pro (top) and DJI Mini 3 Pro (bottom). As we intentionally avoid details from unreliable regions in the design architecture by thresholding to minimize spurious features during reconstruction, pixels associated with sky show the lowest descriptor confidence. By selecting features with top-scored confidence for both detection and descriptor, we find high quality true features while avoiding spurious features for reconstruction.

We compare keypoints and confidence maps from LBurst on captured bursts (top) to R2D2 on the common image (bottom) as shown in \autoref{fig_LB-R2D2}. LBurst outperforms R2D2, detecting more true features and fewer spurious features demonstrating the benefits of using temporal information to handle noise and small motion variations. While increasing the patch size reduces local maxima in the detection confidence map, avoiding excessive spurious features, we set the patch size $M$ as 16 to capture high-quality true features for all methods.

Finally, we evaluate the performance of our LBurst method against both state-of-the-art classical and learning-based methods on captured single images, and other robotic burst-based methods as shown in \autoref{fig_allfeatures}. For a captured scene by DJI Mini 3 Pro, burst-based methods demonstrate an improved capability to locate more true features in both foreground (blue) and background (green) areas. BuFF \cite{aR2022} shows a slightly higher performance in rejecting spurious features, compared to SIFT on burst-merged images \cite{aR2021}. Our feature finder avoids most of the spurious features and regions with low contrast, successfully locating high quality true features associated with the foreground and background within the burst.

Considering state-of-the-art learning-based feature detection techniques on common images of the burst, SuperPoint \cite{detone2018superpoint} exhibits competitive performance for scenes containing buildings but struggles to perform well with images captured by DJI Phantom 4 Pro while R2D2 \cite{revaud2019r2d2} performs poorly compared to other single-image based methods. Considering classical methods, SIFT \cite{lowe2004distinctive} encounters challenges in identifying high-quality features, often yielding more spurious features in comparison to learning and burst-based methods. ORB \cite{Rublee2011} detects an excessive number of features that correspond to substantial noise in the images, failing to represent the low-light scene for reconstruction.

\begin{table*}[t]
\centering
\caption{Average performance of reconstruction for low-light drone-captured burst imagery. Bold: Overall best results, Underline: Best results from our proposed approach.}
\resizebox{\linewidth}{!}{%
\begin{tabular}{cllrrrrrrrrr} 
\hline
 & \multicolumn{2}{c}{\multirow{3}{*}{Methods}} & \multirow{3}{*}{\begin{tabular}[c]{@{}c@{}}Convergence \\rate\end{tabular}} & \multirow{3}{*}{\begin{tabular}[c]{@{}c@{}}Images\\passed \%\end{tabular}} & \multirow{3}{*}{\begin{tabular}[c]{@{}c@{}}Keypoints/\\Image\end{tabular}} & \multirow{3}{*}{\begin{tabular}[c]{@{}c@{}}Putative\\matches/\\Image\end{tabular}} & \multirow{3}{*}{\begin{tabular}[c]{@{}c@{}}Inliers/\\Image\end{tabular}} & \multirow{3}{*}{\begin{tabular}[c]{@{}c@{}}Match\\ratio\end{tabular}} & \multirow{3}{*}{\begin{tabular}[c]{@{}c@{}}Match\\score\end{tabular}} & \multirow{3}{*}{Precision} & \multirow{3}{*}{\begin{tabular}[c]{@{}c@{}}3DPoints/\\Image\end{tabular}} \\
 & \multicolumn{2}{c}{} &  &  &  &  &  &  &  &  &  \\
 & \multicolumn{2}{c}{} &  &  &  &  &  &  &  &  &  \\ 
\cline{2-12}
\multirow{8}{*}{DJI Mini Pro} & \multirow{3}{*}{Conventional} & SIFT & 0.20 & 14.70 & 8202.16 & 83.96 & 71.93 & 0.025 & 0.021 & 0.518 & 10.46 \\
 &  & SuperPoint & \multicolumn{1}{r}{0.20} & \multicolumn{1}{r}{30.70} & \multicolumn{1}{r}{936.10} & \multicolumn{1}{r}{610.30} & \multicolumn{1}{r}{208.10} & \multicolumn{1}{r}{0.599} & \multicolumn{1}{r}{0.296} & \multicolumn{1}{r}{0.439} & \multicolumn{1}{r}{2.59} \\
 &  & R2D2 & \multicolumn{1}{r}{0.20} & \multicolumn{1}{r}{20.00} & \multicolumn{1}{r}{4000.00} & \multicolumn{1}{r}{\textbf{1347.74}} & \multicolumn{1}{r}{310.40} & \multicolumn{1}{r}{0.337} & \multicolumn{1}{r}{0.008} & \multicolumn{1}{r}{0.235} & \multicolumn{1}{r}{1.63} \\
 \cline{2-12}
 & \multirow{3}{*}{Burst with Merge} & SIFT & 0.60 & 45.30 & 6628.68 & 197.51 & 169.41 & 0.061 & 0.052 & 0.805 & 23.06 \\
 &  & SuperPoint & \multicolumn{1}{r}{\textbf{0.80}} & \multicolumn{1}{r}{72.00} & \multicolumn{1}{r}{1044.96} & \multicolumn{1}{r}{728.38} & \multicolumn{1}{r}{\textbf{427.90}} & \multicolumn{1}{r}{\textbf{0.652}} & \multicolumn{1}{r}{\textbf{0.371}} & \multicolumn{1}{r}{0.508} & \multicolumn{1}{r}{20.35} \\
 &  & R2D2 & \multicolumn{1}{r}{\textbf{0.80}} & \multicolumn{1}{r}{76.00} & \multicolumn{1}{r}{4000.00} & \multicolumn{1}{r}{546.30} & \multicolumn{1}{r}{320.90} & \multicolumn{1}{r}{0.137} & \multicolumn{1}{r}{0.080} & \multicolumn{1}{r}{0.508} & \multicolumn{1}{r}{9.86} \\
 \cline{2-12}
 & \multicolumn{2}{l}{BuFF} & \textbf{0.80} & 64.00 & 5603.19 & 177.72 & 158.43 & 0.084 & 0.074 & {\textbf{0.814}} & 25.41 \\
 \cline{2-12}
 & \multicolumn{2}{l}{LBurst (Ours)} & \textbf{\underline{0.80}} & \textbf{\underline{81.30}} & 4000.00 & 690.82 & 351.16 & 0.173 & 0.088 & 0.508 & \textbf{\underline{31.40}} \\ 
\hline
\multirow{8}{*}{DJI Phantom Pro} & \multirow{3}{*}{Conventional} & SIFT & 0.00 & 0.00 & 6116.08 & 5.71 & 2.44 & 0.002 & 0.002 & 0.169 & 0.00 \\
 &  & SuperPoint & \multicolumn{1}{r}{0.40} & \multicolumn{1}{r}{40.00} & \multicolumn{1}{r}{533.28} & \multicolumn{1}{r}{393.28} & \multicolumn{1}{r}{230.50} & \multicolumn{1}{r}{0.701} & \multicolumn{1}{r}{0.386} & \multicolumn{1}{r}{0.523} & \multicolumn{1}{r}{5.13} \\
 &  & R2D2 & \multicolumn{1}{r}{0.00} & \multicolumn{1}{r}{0.00} & \multicolumn{1}{r}{4000.00} & \multicolumn{1}{r}{1025.57} & \multicolumn{1}{r}{103.52} & \multicolumn{1}{r}{0.256} & \multicolumn{1}{r}{0.026} & \multicolumn{1}{r}{0.101} & \multicolumn{1}{r}{0.00} \\
 \cline{2-12}
 & \multirow{3}{*}{Burst with Merge} & SIFT & 0.00 & 0.00 & 2218.68 & 27.91 & 17.80 & 0.013 & 0.008 & 0.580 & 0.00 \\
 &  & SuperPoint & \multicolumn{1}{r}{0.40} & \multicolumn{1}{r}{40.00} & \multicolumn{1}{r}{495.84} & \multicolumn{1}{r}{362.28} & \multicolumn{1}{r}{219.20} & \multicolumn{1}{r}{ \textbf{0.704}} & \multicolumn{1}{r}{ \textbf{0.406}} & \multicolumn{1}{r}{0.564} & \multicolumn{1}{r}{6.35} \\
 &  & R2D2 & \multicolumn{1}{r}{0.40} & \multicolumn{1}{r}{40.00} & \multicolumn{1}{r}{4000.00} & \multicolumn{1}{r}{\textbf{1346.90}} & \multicolumn{1}{r}{420.68} & \multicolumn{1}{r}{0.337} & \multicolumn{1}{r}{0.105} & \multicolumn{1}{r}{0.300} & \multicolumn{1}{r}{5.40} \\
 \cline{2-12}
 & \multicolumn{2}{l}{BuFF} & 0.40 & 28.00 & 1483.42 & 17.74 & 14.71 & 0.032 & 0.027 & \textbf{0.859} & 11.78 \\
 \cline{2-12}
 & \multicolumn{2}{l}{LBurst (Ours)} & \textbf{\underline{1.00}} & \textbf{\underline{89.30}} & 4000.00 & 1179.38 & \textbf{\underline{480.42}} & 0.295 & 0.120 & 0.432 & \textbf{\underline{14.85}} \\
\hline
\end{tabular}
}
\label{table_reconstruction}
\end{table*}
\subsubsection{Reconstruction Performance}
Following the feature comparison approach in~\cite{Schonberger2017}, we evaluate the reconstruction performance of the drone-captured imagery, as in \autoref{table_reconstruction}. We compare convergence rate, defined as the number of reconstructed scenes refined by bundle adjustment, and image pass, which refers to the number of images used for reconstruction from the input images. Additionally, we compare the count of keypoints, the number of putative feature matches, the quantity of matches classified as inliers (the proportion of detected features yielding putative matches), precision (the proportion of putative matches yielding inlier matches), matching score (the proportion of detected features yielding inlier matches), and the average number of 3D points per reconstructed image. Importantly, our evaluation considers more challenging images, while other methods focus on a smaller, less challenging subset, highlighting the completeness of our models. In the table, bold highlights the best overall results and underlined results are the best results from our proposed method.


We compute 4000 keypoints, employing detection and descriptor thresholds set at 0.7 for both our approach and R2D2. We evaluate reconstruction performance using single-image and burst-merged image strategies by operating classical SIFT, and learning-based methods including R2D2 and SuperPoint. As we employ burst directly to find features for our method, we also compare against physics-based burst feature finder. 

Our method outperforms state-of-the-art feature detection methods on both conventional and burst-merged images, by reconstructing the majority of scenes captured by the DJI Mini Pro 3 with the highest number of 3D points. Compared to physics-based burst feature finder, our method detects twice the inlier matches. For images captured by the DJI Phantom 4 Pro, our method reconstructs all scenes with the highest number of inlier matches and 3D points, resulting in more complete 3D sparse models. While SuperPoint takes advantage of the SNR boost associated with burst-merged images and demonstrates competitive performance in detecting inlier matches, it struggles to identify 3D points. However, R2D2 with a design that encourages AP optimization, similar to our approach, tends to generate an excessive number of spurious putative matches. Overall, our method benefits from improved feature performance, which leads to a more complete 3D reconstruction in low light with an overall highest convergence rate.


\subsection{Speed}
We repeatedly run the network by varying scales to compute scale-invariant features for reconstruction. We maintain the largest dimension at $2^{10}$ pixels and downsample to $2^8$ at intervals of $2^{1/4}$. Features are located by identifying local maxima at each scale with a patch size of 16. It takes an average of 2.27 seconds on an NVIDIA GeForce RTX 3080 Laptop GPU to extract keypoints at all scales for a 5-frame robotic burst, comparable to the time R2D2 takes for a single image during inference. Our method, however, introduces an additional overhead for loading more data and takes 1.34 minutes for a complete reconstruction using COLMAP.

\section{Discussion}
\label{sec:discussion}
The number of keypoints depends on the captured scenes, and exceeding a threshold can result in spurious features, especially in low-contrast, noisy regions. Similar to R2D2, the performance of LBurst can be optimized by adjusting parameters such as patch size, detector and descriptor thresholds, and scale levels. Using more images in a burst and fine-tuning the model with captured drone-burst data could further improve feature detection and description. Our method is also influenced by domain-specific training, which may limit broader applicability, and adjustments are needed for drone-based, bird's-eye perspective applications.
\section{Conclusions}
\label{sec:conclusions}
We developed a learned burst feature finder that detects and describes features directly within robotic bursts for light-constrained 3D reconstruction. By leveraging temporal information in low SNR images, our method computes noise-tolerant features even in challenging millilux environments. We demonstrated successful reconstruction with decreased failure cases due to non-convergence compared to conventional and burst-based methods. Furthermore, we showed accurate matches yielding complete models with a higher number of 3D points. For future work, we anticipate incorporating multi-scale training and IMU integration to further improve reconstruction in visually challenging conditions.

{\small
	\bibliographystyle{IEEEtran}
	\bibliography{./references}

\begin{thebibliography}{10}
\providecommand{\url}[1]{#1}
\csname url@rmstyle\endcsname
\providecommand{\newblock}{\relax}
\providecommand{\bibinfo}[2]{#2}
\providecommand\BIBentrySTDinterwordspacing{\spaceskip=0pt\relax}
\providecommand\BIBentryALTinterwordstretchfactor{4}
\providecommand\BIBentryALTinterwordspacing{\spaceskip=\fontdimen2\font plus
\BIBentryALTinterwordstretchfactor\fontdimen3\font minus
  \fontdimen4\font\relax}
\providecommand\BIBforeignlanguage[2]{{%
\expandafter\ifx\csname l@#1\endcsname\relax
\typeout{** WARNING: IEEEtran.bst: No hyphenation pattern has been}%
\typeout{** loaded for the language `#1'. Using the pattern for}%
\typeout{** the default language instead.}%
\else
\language=\csname l@#1\endcsname
\fi
#2}}

\bibitem{mehta2021domain}
A.~Mehta, H.~Sinha, M.~Mandal, and P.~Narang, ``Domain-aware unsupervised
  hyperspectral reconstruction for aerial image dehazing,'' in \emph{Winter
  Conference on Applications of Computer Vision (WACV)}, 2021, pp. 413--422.

\bibitem{zhu2021revisiting}
S.~Zhu, T.~Yang, and C.~Chen, ``Revisiting street-to-aerial view image
  geo-localization and orientation estimation,'' in \emph{Winter Conference on
  Applications of Computer Vision (WACV)}, 2021, pp. 756--765.

\bibitem{ashraf2021dogfight}
M.~W. Ashraf, W.~Sultani, and M.~Shah, ``Dogfight: Detecting drones from drones
  videos,'' in \emph{Computer Vision and Pattern Recognition (CVPR)}, 2021, pp.
  7067--7076.

\bibitem{asanomi2023multi}
T.~Asanomi, K.~Nishimura, and R.~Bise, ``Multi-frame attention with
  feature-level warping for drone crowd tracking,'' in \emph{Winter Conference
  on Applications of Computer Vision (WACV)}, 2023, pp. 1664--1673.

\bibitem{zhu2021msnet}
X.~Zhu, J.~Liang, and A.~Hauptmann, ``{MSNet: A multilevel instance
  segmentation network for natural disaster damage assessment in aerial
  videos},'' in \emph{Winter Conference on Applications of Computer Vision
  (WACV)}, 2021, pp. 2023--2032.

\bibitem{hasinoff2016burst}
S.~W. Hasinoff, D.~Sharlet, R.~Geiss, A.~Adams, J.~T. Barron, F.~Kainz,
  J.~Chen, and M.~Levoy, ``Burst photography for high dynamic range and
  low-light imaging on mobile cameras,'' \emph{{ACM Transactions on Graphics
  (TOG)}}, vol.~35, no.~6, pp. 1--12, 2016.

\bibitem{aR2021}
A.~Ravendran, M.~Bryson, and D.~G. Dansereau, ``{Burst imaging for
  light-constrained structure-from-motion},'' \emph{Robotics and Automation
  Letters (RA-L)}, vol.~7, no.~2, pp. 1040--1047, 2021.

\bibitem{aR2022}
------, ``{BuFF: Burst feature finder for light-constrained 3D
  reconstruction},'' \emph{Robotics and Automation Letters (RA-L)}, vol.~8,
  no.~12, pp. 8438--8445, 2023.

\bibitem{detone2018superpoint}
D.~DeTone, T.~Malisiewicz, and A.~Rabinovich, ``{SuperPoint: Self-supervised
  interest point detection and description},'' in \emph{Computer Vision and
  Pattern Recognition (CVPR) Workshops}, 2018, pp. 224--236.

\bibitem{revaud2019r2d2}
J.~Revaud, C.~De~Souza, M.~Humenberger, and P.~Weinzaepfel, ``{R2D2: Reliable
  and repeatable detector and descriptor},'' \emph{Advances in Neural
  Information Processing Systems (NeurIPS)}, vol.~32, 2019.

\bibitem{he2018local}
K.~He, Y.~Lu, and S.~Sclaroff, ``Local descriptors optimized for average
  precision,'' in \emph{Computer Vision and Pattern Recognition (CVPR)}, 2018,
  pp. 596--605.

\bibitem{balntas2017hpatches}
V.~Balntas, K.~Lenc, A.~Vedaldi, and K.~Mikolajczyk, ``{HPatches: A benchmark
  and evaluation of handcrafted and learned local descriptors},'' in
  \emph{Computer Vision and Pattern Recognition (CVPR)}, 2017, pp. 5173--5182.

\bibitem{schonberger2016structure}
J.~L. Schonberger and J.-M. Frahm, ``Structure-from-motion revisited,'' in
  \emph{Computer Vision and Pattern Recognition (CVPR)}, 2016, pp. 4104--4113.

\bibitem{liba2019handheld}
O.~Liba, K.~Murthy, Y.-T. Tsai, T.~Brooks, T.~Xue, N.~Karnad, Q.~He, J.~T.
  Barron, D.~Sharlet, R.~Geiss, \emph{et~al.}, ``Handheld mobile photography in
  very low light,'' \emph{{ACM Transactions on Graphics (TOG)}}, vol.~38,
  no.~6, pp. 1--16, 2019.

\bibitem{wronski2019handheld}
B.~Wronski, I.~Garcia-Dorado, M.~Ernst, D.~Kelly, M.~Krainin, C.-K. Liang,
  M.~Levoy, and P.~Milanfar, ``Handheld multi-frame super-resolution,''
  \emph{{ACM Transactions on Graphics (TOG)}}, vol.~38, no.~4, pp. 1--18, 2019.

\bibitem{mildenhall2018burst}
B.~Mildenhall, J.~T. Barron, J.~Chen, D.~Sharlet, R.~Ng, and R.~Carroll,
  ``Burst denoising with kernel prediction networks,'' in \emph{{Computer
  Vision and Pattern Recognition (CVPR)}}, 2018, pp. 2502--2510.

\bibitem{tyszkiewicz2020disk}
M.~Tyszkiewicz, P.~Fua, and E.~Trulls, ``Disk: Learning local features with
  policy gradient,'' \emph{Advances in Neural Information Processing Systems},
  vol.~33, pp. 14\,254--14\,265, 2020.

\bibitem{han2015matchnet}
X.~Han, T.~Leung, Y.~Jia, R.~Sukthankar, and A.~C. Berg, ``{MatchNet: Unifying
  feature and metric learning for patch-based matching},'' in \emph{Computer
  Vision and Pattern Recognition (CVPR)}, 2015, pp. 3279--3286.

\bibitem{yi2016lift}
K.~M. Yi, E.~Trulls, V.~Lepetit, and P.~Fua, ``{LIFT: Learned invariant feature
  transform},'' in \emph{European Conference on Computer Vision (ECCV)}, 2016,
  pp. 467--483.

\bibitem{dusmanu2019d2net}
M.~Dusmanu, I.~Rocco, T.~Pajdla, M.~Pollefeys, J.~Sivic, A.~Torii, and
  T.~Sattler, ``{D2-Net: A trainable CNN for joint description and detection of
  local features},'' in \emph{{Computer Vision and Pattern Recognition
  (CVPR)}}, 2019, pp. 8092--8101.

\bibitem{lowe2004distinctive}
D.~G. Lowe, ``Distinctive image features from scale-invariant keypoints,''
  \emph{{International Journal of Computer Vision (IJCV)}}, vol.~60, no.~2, pp.
  91--110, 2004.

\bibitem{Bay2006}
H.~Bay, T.~Tuytelaars, and L.~V. Gool, ``{SURF: Speeded up robust features},''
  in \emph{European Conference on Computer Vision (ECCV)}.\hskip 1em plus 0.5em
  minus 0.4em\relax Springer, 2006, pp. 404--417.

\bibitem{Rublee2011}
E.~Rublee, V.~Rabaud, K.~Konolige, and G.~Bradski, ``{ORB: An efficient
  alternative to SIFT or SURF},'' in \emph{International Conference on Computer
  Vision (ICCV)}, 2011, pp. 2564--2571.

\bibitem{lecun1995convolutional}
Y.~LeCun, Y.~Bengio, \emph{et~al.}, ``Convolutional networks for images,
  speech, and time series,'' \emph{The handbook of brain theory and neural
  networks}, vol. 3361, no.~10, p. 1995, 1995.

\bibitem{wang2023attention}
C.~Wang, R.~Xu, K.~Lu, S.~Xu, W.~Meng, Y.~Zhang, B.~Fan, and X.~Zhang,
  ``Attention weighted local descriptors,'' \emph{IEEE Transactions on Pattern
  Analysis and Machine Intelligence}, vol.~45, no.~9, pp. 10\,632--10\,649,
  2023.

\bibitem{tian2017l2}
Y.~Tian, B.~Fan, and F.~Wu, ``{L2-Net: Deep learning of discriminative patch
  descriptor in Euclidean space},'' in \emph{Computer Vision and Pattern
  Recognition (CVPR)}, 2017, pp. 661--669.

\bibitem{sattler2018benchmarking}
T.~Sattler, W.~Maddern, C.~Toft, A.~Torii, L.~Hammarstrand, E.~Stenborg,
  D.~Safari, M.~Okutomi, M.~Pollefeys, J.~Sivic, \emph{et~al.}, ``{Benchmarking
  6DOF outdoor visual localization in changing conditions},'' in \emph{Computer
  Vision and Pattern Recognition (CVPR)}, 2018, pp. 8601--8610.

\bibitem{radenovic2018revisiting}
F.~Radenovi{\'c}, A.~Iscen, G.~Tolias, Y.~Avrithis, and O.~Chum, ``{Revisiting
  Oxford and Paris: Large-scale image retrieval benchmarking},'' in
  \emph{Computer Vision and Pattern Recognition (CVPR)}, 2018, pp. 5706--5715.

\bibitem{kingma2014adam}
D.~P. Kingma and J.~Ba, ``{Adam: A method for stochastic optimization},''
  \emph{aXiv preprint arXiv:1412.6980}, 2014.

\bibitem{Schonberger2017}
J.~L. Schonberger, H.~Hardmeier, T.~Sattler, and M.~Pollefeys, ``Comparative
  evaluation of hand-crafted and learned local features,'' in \emph{Computer
  Vision and Pattern Recognition (CVPR)}, 2017, pp. 1482--1491.

\end{thebibliography}
}
\end{document}